\documentclass[lettersize,journal]{IEEEtran}
\usepackage{amsmath,amsfonts}
\usepackage{algorithmic}
\usepackage{algorithm}
\usepackage{array}
\usepackage{adjustbox}
\usepackage{bm}

\usepackage[caption=false,font=normalsize,labelfont=sf,textfont=sf]{subfig}
\usepackage{textcomp}
\usepackage{stfloats}
\usepackage{url}
\usepackage{verbatim}
\usepackage{graphicx}
\usepackage{booktabs}
\usepackage{multirow}
\usepackage{color}
\usepackage[colorlinks = true,
            linkcolor = blue,
            urlcolor  = blue,
            citecolor = blue,
            anchorcolor = blue]{hyperref}
\usepackage{xcolor}

\begin{document}

\title{Spatial-Temporal Mixture-of-Graph-Experts for Multi-Type Crime Prediction}

\author{Ziyang Wu, Fan Liu, Jindong Han, \IEEEmembership{member,~IEEE}, Yuxuan Liang, Hao LIU,~\IEEEmembership{Senior member,~IEEE,}

\thanks{Manuscript received xx xx, xxxx; revised xx xx, xxxx.}
\thanks{Ziyang Wu adn Fan Liu are with the Thrust of Artificial Intelligence, Hong Kong University of Science and Technology (Guangzhou), Guangzhou 511458, China. E-mail: zwu390@connect.hkust-gz.edu.cn; fliu236@connect.hkust-gz.edu.cn.}
\thanks{Jindong Han is with the Division of Emerging Interdisciplinary Areas, The Hong Kong University of Science and Technology, Hong Kong SAR, China (e-mail: jhanao@connect.ust.hk).}
\thanks{Yuxuan Liang is with the Thrust of Intelligent Transportation \& the Thrust of Data Science and Analytics, Hong Kong
University of Science and Technology (Guangzhou), Guangzhou 511458, China. E-mail: yuxuanliang@hkust-gz.edu.cn.}
\thanks{Hao Liu is with the Thrust of Artificial Intelligence, Hong Kong University of Science and Technology (Guangzhou), Guangzhou 511458, China,
and also with the Department of Computer Science and Engineering,
Hong Kong University of Science and Technology, Hong Kong SAR,
China. E-mail: liuh@ust.hk.}}

 \markboth{ IEEE TRANSACTIONS ON KNOWLEDGE AND DATA ENGINEERING,~Vol.~xx, No.~x, xx~xxxx}
{Shell \MakeLowercase{\textit{et al.}}: A Sample Article Using IEEEtran.cls for IEEE Journals}

\IEEEpubid{\begin{minipage}{\textwidth}\ \centering
		1041-4347 \copyright 2024 IEEE. Personal use is permitted, but republication/redistribution requires IEEE permission. \\
See https://www.ieee.org/publications/rights/index.html for more information.
\end{minipage}}


\maketitle

\begin{abstract}

As various types of crime continue to threaten public safety and economic development, predicting the occurrence of multiple types of crimes becomes increasingly vital for effective prevention measures. Although extensive efforts have been made, most of them overlook the heterogeneity of different crime categories and fail to address the issue of imbalanced spatial distribution. In this work, we propose a Spatial-Temporal Mixture-of-Graph-Experts (ST-MoGE) framework for collective multiple-type crime prediction. To enhance the model's ability to identify diverse spatial-temporal dependencies and mitigate potential conflicts caused by spatial-temporal heterogeneity of different crime categories, we introduce an attentive-gated Mixture-of-Graph-Experts (MGEs) module to capture the distinctive and shared crime patterns of each crime category. 
Then, we propose Cross-Expert Contrastive Learning(CECL) to update the MGEs and force each expert to focus on specific pattern modeling, thereby reducing blending and redundancy. Furthermore, to address the issue of imbalanced spatial distribution, we propose a Hierarchical Adaptive Loss Re-weighting (HALR) approach to eliminate biases and insufficient learning of data-scarce regions. To evaluate the effectiveness of our methods, we conduct comprehensive experiments on two real-world crime datasets and compare our results with twelve advanced baselines. The experimental results demonstrate the superiority of our methods.
\end{abstract}

 
\section{Introduction}
 \IEEEPARstart{A}{s} the incidence of various crimes (e.g., theft, assault, robbery, etc.) continues to rise, they have become a significant threat to public safety and economic development~\cite{wortley2016environmental}. Contrary to the intuitive belief that criminal acts are random and unpredictable, crime pattern theory reveals a different reality, suggesting that crimes are either planned or opportunistic, following implicit patterns~\cite{zhao2018crime}. The goal of crime prediction, therefore, endeavors to unravel these patterns, as accurate forecasting of crime occurrences stands as a cornerstone for effective crime prevention. By empowering policymakers and law enforcement agencies to adopt proactive strategies, crime prediction plays a pivotal role in shaping urban governance and enhancing public safety.

Due to its importance, numerous machine learning based crime prediction methods have recently been developed, generally categorized into deep attentive-based and graph-based approaches. The deep attentive-based approach ~\cite{xia2021spatial, yao2019revisiting} aims to model crime dynamics using various attention mechanisms. For example, DeepCrime models crime dynamics using external feature adaptive fusion. On the other hand, graph-based approaches ~\cite{yu2018spatio, bai2020adaptive, zheng2020gman, li2022spatial} aim to capture the spatial-temporal relationships among different crime patterns.  For instance, Wu et al. ~\cite{wu2020hierarchically} explore the use of spatial-temporal Graph Neural Networks (GNNs) for crime prediction, leveraging the spatial modeling capabilities of GNNs. Despite these significant efforts, these methods still struggle to effectively model the spatial-temporal dependencies among different crime categories due to their spatial-temporal heterogeneity.

\begin{figure}[!t]
    \centering
    \includegraphics[width=1.0\columnwidth]{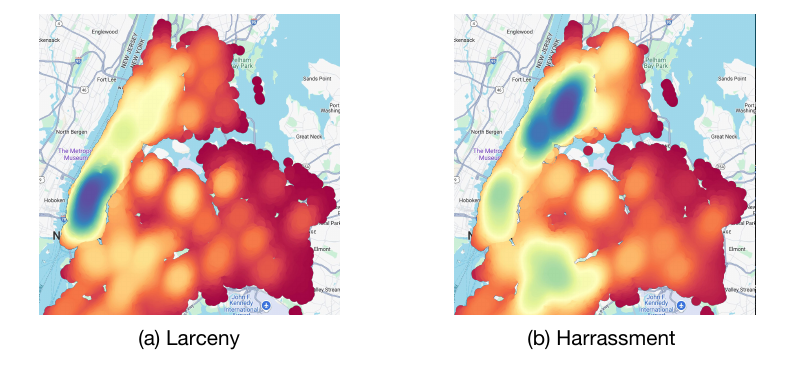}
    \caption{Spatial Distribution of Crime Occurrences in New York City}
    \label{fig1}
\end{figure}

The spatial-temporal heterogeneity among different crime categories is primarily driven by the distinct spatial-temporal patterns characteristic of each crime category. Crime activities in each category typically reflect unique patterns. For instance, burglaries are intuitively more likely to occur in residential zones than in commercial or public areas, whereas larcenies are more prevalent in the latter. As depicted in Figure \ref{fig1}, different crime categories exhibit significant disparities in their spatial distributions, highlighting the pronounced pattern heterogeneity among them. When traditional spatial-temporal prediction approaches are applied to crime prediction, they are often constrained by their model architecture, which typically only captures shared spatial-temporal dependencies across all crime categories. Such approach overlooks the distinctive patterns unique to each category, which can lead to two key issues: 1) The unique crime patterns, especially those that are distinctive, cannot be fully captured, resulting in incomplete pattern modeling; 2) Different categories of crime patterns can be mutually contradictory in specific regions, complicating accurate pattern modeling. However, simply introducing a more tailored model architecture or increasing the model size is insufficient to address this pattern heterogeneity. Therefore, a fundamental question arises: \textit{ How can we develop a prediction model that fully captures the heterogeneous crime patterns while alleviating the conflicts among categories?}

\IEEEpubidadjcol

\begin{figure}[!t]
    \centering
    \includegraphics[width=1.0\columnwidth]{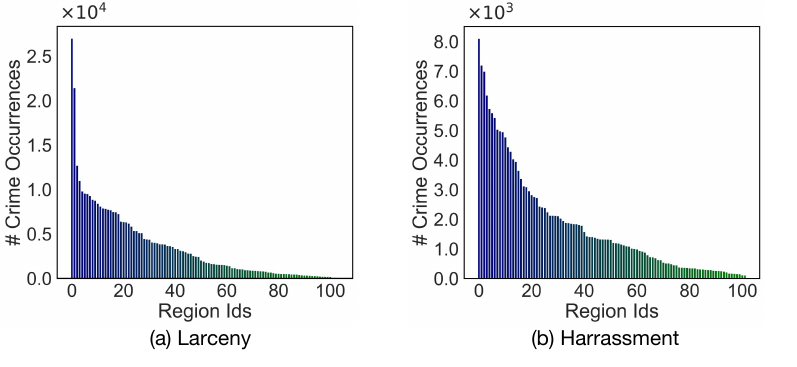}
    \caption{Amounts of Regional Crime Occurrence in New York City}
    \label{fig2}
\end{figure}

Recently, the Mixture-of-Experts (MoE) architecture has demonstrated its superiority in handling heterogeneous data due to its remarkable adaptability and generalization capabilities~\cite{artetxe2022efficient, riquelme2021scaling}. However, directly adopting the MoE architecture for the multi-type crime prediction presents a non-trivial task. First, there is the issue of dynamically routing different experts to integrate the extracted spatial-temporal knowledge. Most existing MoE methods are designed to handle static tasks, whereas crime prediction is typically a dynamic task. This dynamic nature arises from the fact that crime patterns are highly variable across different regions and time periods. For instance, property crimes might peak in certain residential areas during nighttime, while violent crimes may be more prevalent in commercial areas during weekends. As a result, the system must route the specific experts for different crime categories, adapting to changes in spatial and temporal crime distributions, which is challenging. Second, the complexity of dynamically modeling the various crime patterns adds another layer of difficulty. Crime data naturally exhibits spatially imbalanced distributions. As shown in Figure~\ref{fig2}, different crime categories are unevenly distributed across various regions of New York City. Traditional MoE frameworks can be easily influenced by regions with extreme numbers of crime occurrences, potentially compromising prediction performance in other regions. Therefore, avoiding biased modeling caused by imbalanced spatial distribution presents another significant challenge. These challenges necessitate a sophisticated mechanism to ensure that the MoE framework can adapt to the dynamic and heterogeneous nature of crime data.

To this end, we propose a new spatio-temporal crime prediction framework, named Spatial-Temporal Mixture-of-Graph-Experts (ST-MoGE), designed to address the spatial-temporal heterogeneity posed by different crime categories. Specifically, we introduce the Attentive-gated Mixture-of-Graph-Experts (MGEs) module. This module comprises multiple spatial-temporal graph learning experts—both crime-specific and universal experts—to fully capture the heterogeneous and universal crime patterns. An attentive gating mechanism is employed to dynamically route specific experts for the selective integration of the extracted spatial-temporal knowledge from each expert. Additionally, to ensure that the optimal crime-dependent experts are routed and to decompose the learning of unique crime patterns in parallel, we propose cross-expert contrastive learning (CECL) to ensure that each crime-dependent expert focuses on learning its specific crime pattern. Furthermore, to address the challenge of imbalanced spatial distribution, we propose regional-aware predictors. These predictors dynamically adjust weights across different regions to mitigate disproportionate influences during model training.

The key contributions can be summarized as follows:

\textbullet  We present ST-MoGE, a tailored framework for collective multi-type crime prediction an innovative framework for crime prediction. To the best of our knowledge, this study is the first attempt to address the spatial-temporal heterogeneity of different crime categories by MOE architecture. 

\textbullet We propose an MoE-architectured spatial-temporal network that comprehensively captures heterogeneous spatial-temporal characteristics posed by diverse crime categories. We further devise a contrastive learning approach, relieving the redundancy and enhancing the concentration of each expert. Moreover, we devise a hierarchical adaptive loss-reweighting algorithm to relieve the training biases caused by the imbalanced spatial distribution of crimes.

\textbullet We conduct comprehensive experiments on two real-world crime datasets from New York City and Chicago, evaluating 12 spatial-temporal prediction methods from various angles. The experimental results demonstrate that our ST-MoGE significantly outperforms existing methods.

\section{Related Works}
\label{sec_related}
\subsection{Spatial-Temporal Prediction}

Spatial-temporal prediction involves analyzing and predicting the evolution of state or value across both spatial and temporal dimensions. It has been widely used for tasks of urban computing, such as traffic flow prediction~\cite{liu2022practical} and public transportation optimization~\cite{liu2020polestar}. Extensive methodologies are proposed for spatial-temporal prediction. Early methodologies involve employing statistical or conventional machine learning models, including ARIMA~\cite{box2015time}, Random Forest Regression~\cite{yang2019understanding}, SVR~\cite{wu2004travel} and KNN~\cite{cai2016spatiotemporal}. In recent years, deep learning methods, benefiting from their capacity for capturing complex spatial-temporal dependencies, have shown remarkable performance in various urban computing tasks involving spatial-temporal prediction~\cite{luo2020spatial,zhang2021mugrep,he2018profiling}. Time-series prediction methods like Recurrent Neural Networks (RNNs) and Temporal Convolutional Networks (TCNs) are widely applied for modeling temporal dependencies. RNNs are used in models such as D-LSTM~\cite{yu2017deep}, ST-RNN \cite{wang2017predrnn}. On the other hand, TCNs are leveraged in GWN~\cite{wu2019graph} and MTGNN~\cite{wu2020connecting}, demonstrating their effectiveness in handling temporal dynamics. To capture spatial dependencies, convolutional neural networks (CNNs) are often applied to grid-structured data~\cite{li2022spatial}. More recently, graph neural networks (GNNs) have been adopted to model spatial dependencies in non-Euclidean structured spatial-temporal data. For instance, several methods leverage spectral graphs, such as STGCN~\cite{yu2018spatio} and DCRNN~\cite{li2018diffusion}. To account for dynamic spatial dependencies, graph attention neural networks have been introduced to dynamically calculate region-specific importance, as seen in GMAN \cite{wu2020connecting}. These models do not require predefined spatial information, allowing them to discover spatial correlations in a data-driven manner.  However, the aforementioned methods could be limited when applied to crime prediction due to several imbalanced data distributions and issues with multi-objective prediction. 

\subsection{Crime Prediction} 
Crime prediction has been widely researched over the years. Initially, statistical models and conventional machine learning methods were employed, including ARIMA~\cite{catlett2018data, alves2018crime}, KNN~\cite{shermila2018crime, kumar2020crime}, Random Forest~\cite{raza2021data, yao2020prediction}, Decision Tree~\cite{ma2022eadtc, zhang2022interpretable}, and XGBoost~\cite{almuhanna2021prediction, yan2022research}. In recent years, deep learning-based methods have also gained significant attention in crime prediction~\cite{chun2019crime, yi2018integrated}.

Some researchers use external information to boost prediction performance~\cite{yang2018crimetelescope, saraiva2022crime, shah2021crime, boukabous2022multimodal}. For example,~\cite{wang2016crime} estimates regional crime rates using point-of-interest (POI) data and demographic information. Huang et al.~\cite{huang2018deepcrime} developed DeepCrime, which builds static region embeddings with POI data to capture regional traits and uses hierarchical GRUs for prediction. AttenCrime~\cite{zhao2023classification} incorporates external spatial-temporal data like anomaly data and taxi data, employing attention mechanisms to model these cross-domain relationships.

Data sparsity is a key challenge in crime prediction, limiting the model's ability to detect patterns in areas with few crime occurrences. Zhao et al.~\cite{zhao2017exploring} tackled this by using transfer learning, extracting crime patterns from areas with abundant data and applying them to regions with low crime rates. Similarly, in~\cite{zhou2022unsupervised}, an unsupervised domain adaptation method was introduced for cross-city crime risk prediction. Li et al.~\cite{li2022spatial} also used self-supervised learning for data augmentation in ST-HSL, incorporating a hypergraph convolution network for crime pattern modeling. Fine-grained crime prediction worsens data sparsity due to the increased detail, making the problem even more significant~\cite{wu2020hierarchically,zhao2023classification,liang2022towards}. In STtrans~\cite{wu2020hierarchically}, a shared embedding space for pattern extraction and adversarial training helped address this issue. Zhao et al.~\cite{zhao2023classification} also proposed a classification-labeled continuousization strategy to convert sparse crime data into continuous signals, easing the sparsity problem in fine-grained predictions.

Unlike deep learning methods, some studies use mathematical models to predict crime patterns, offering faster inference and better explainability~\cite{zhao2017modeling, zhao2022multi, agarwal2018crime, alves2018crime}. For instance, Zhao et al.~\cite{zhao2017modeling} introduced a tensor factorization approach for single-category crime prediction. In their later work~\cite{zhao2022multi}, they expanded to multi-category crime prediction, designing a mathematical model to capture dependencies across spatial, temporal, and categorical dimensions.

\subsection{Mixture-of-Expert}
 Mixture-of-Expert (MoE) indicates a machine learning architecture that combines multiple specialized sub-models, each trained to handle different subsets of data or tasks, to improve overall performance. In recent years, MoE has been applied in various domains (e.g., computer visions~\cite{park2018megan, riquelme2021scaling}, nature language processes~\cite{xue2024openmoe,du2022glam}), presenting remarkable effectiveness in handling heterogeneous data and complex tasks. Several existing studies also adopt MoE for spatial-temporal prediction. For instance, GESME-Net~\cite{rahman2020gated} introduces a spatial-temporal MoE network with CNNs and RNNs as experts, applied to handle the source heterogeneity posed by multi-city ride-hailing demand prediction. Liu et. al.~\cite{liu2023st} propose ST-MoE, which constructs embeddings encoded with spatial-temporal knowledge and adopts MLPs as experts for prediction.  CP-MoE~\cite{jiang2024interpretable} employs multiple tailored adaptive graph learners as experts to capture traffic congestion spatial-temporal patterns from various aspects and further introduces specialized experts to identify stable trends and periodic patterns from traffic data. An ordinal regression strategy is further adapted to facilitate effective collaboration among different experts.

\section{Preliminaries and Problem Definition}
\label{sec_pre}
In this section, we first introduce several definitions in this paper and then formally formulate the research problem.

\textit{Definition 1 (Region Graph $\mathcal{G}$)}. In this study, we represent the topological structure of a city as a graph $\mathcal{G}=(\mathcal{V},\mathcal{E},\mathcal{A})$. The city is equally divided into $N$ regions using, with each region labeled as $R = \{r_i|r_i \in r_1,...,r_n,...,r_V\}$. Each region serves as a node in $\mathcal{E}$. The set $\mathcal{V}$ represents a collection of edges. If two regions are adjacent geographically, their corresponding nodes in the graph are considered connected. $\mathcal{A}$ represents the adjacency matrix, defined as $\mathcal{A} \in \mathbb{R}^{N \times N}$, where

\begin{equation}
    \mathcal{A}_{i,j}=\left\{
             \begin{array}{l}
             1, \text{ if } r_i \text{ and } r_j \text{ are connected} \\
             0, \text{ else} .  
             \end{array}
    \right.
\end{equation}

\textit{Definition 2 (Crime Tensor)}. The raw crime data comprises reports of crime occurrences. Each report includes details such as the category of crime, precise timestamp, and geographical coordinates. By associating each crime occurrence with its respective time slot and region, we construct the crime tensor $\bm{X} \in \mathbb{R}^{N \times T \times C}$, where $N$, $T$, and $C$ represent the number of regions, time slots, and crime categories, respectively. Within the crime tensor, each entry $\bm{X}_{n,t,c}$ denotes the occurrence of a crime in category $c$ within region $r$ during time slot $t$.

\textit{Problem Formalization} 
Given the historical crime tensor $\bm{X} \in \mathbb{R}^{N \times T \times C}$ from previous $T$ time slots and the region graph $\mathcal{G}$, the crime prediction problem aims to learn a mapping function $\mathcal{F}(\cdot)$ to predict the crime occurrence of each category in each region during the next time slot $T+1$, 
 \begin{equation}
    \mathcal{F}:(\bm{X}_T,\mathcal{G})\mapsto\hat{\bm{X}}_{T+1}\in\mathbb{R}^{N \times C},
\end{equation}
where $\hat{\bm{X}}_{T+1}$ represents the estimated crime tensor in next time slot $T+1$, and considered as the prediction results.
\section{Methodology}
\label{sec_met}

\begin{figure*}
    \centering
    \includegraphics[width=1.0\linewidth]{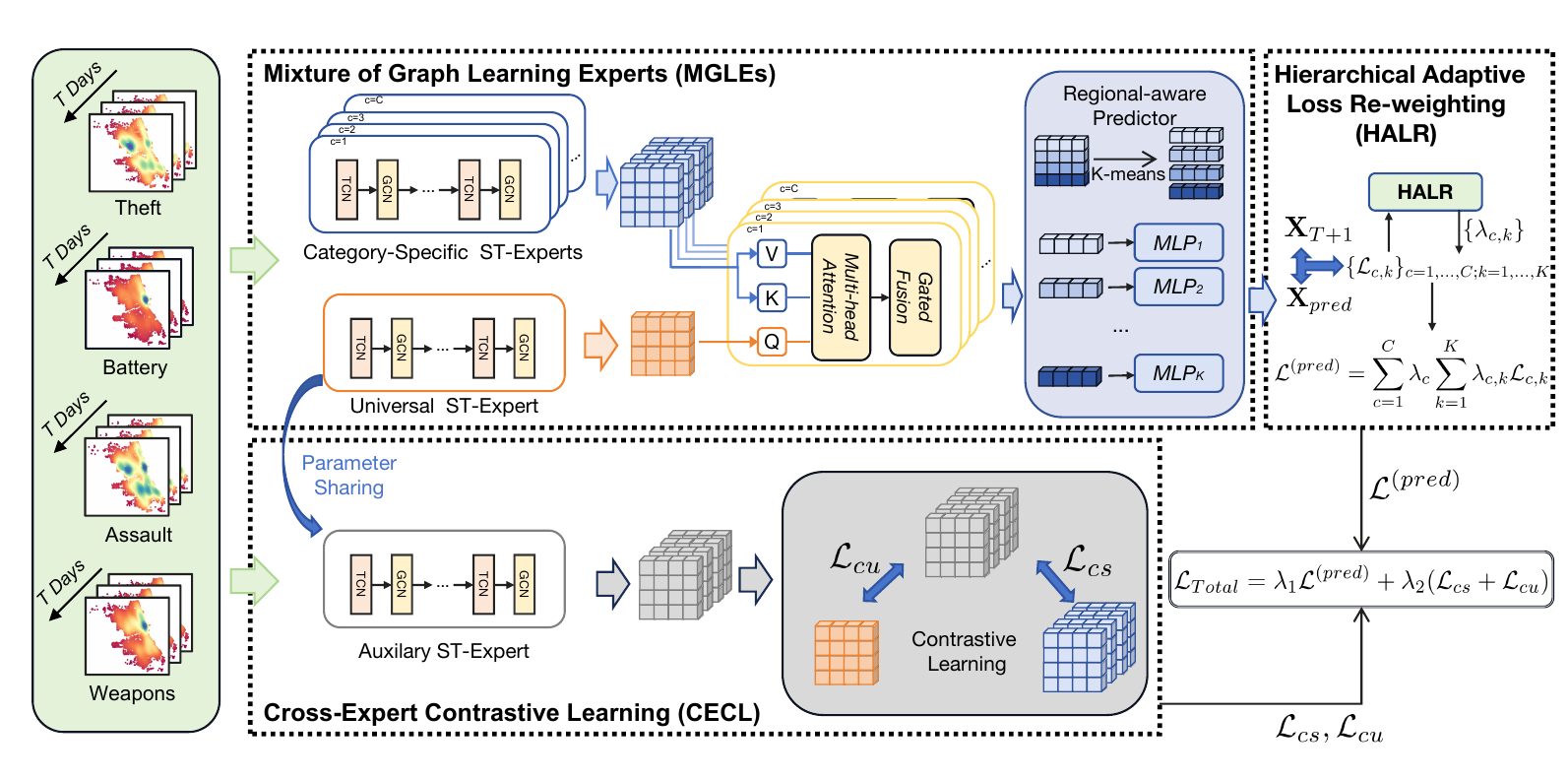}
    \caption{The Architecture of the Proposed ST-MoGE Framework.  }
    \label{fig_model}
\end{figure*}

Figure \ref{fig_model} presents an overview of the ST-MoGE framework, showing the architecture of the \textit{Attentive-Gated Mixture-of-graph-Experts (MGEs)}. It contains multiple spatial-temporal graph learning experts (ST-expert), including the category-specific ST-experts and the universal ST-expert, to capture the shared and category-specific crime patterns, respectively.  Spatial attentive gates are further introduced to selectively route the shared spatial-temporal knowledge of each region for different crime categories. The regional-aware predictor is further applied, in which the regions are clustered based on their spatial-temporal characteristics, and tailored predictors are deployed to obtain the prediction results.

To enhance the prediction capacity of MGEs module, two training strategies are further applied: (1) \textit{Cross-Expert Contrastive Learning (CECL)}: It employs an auxiliary expert shared parameter with the universal expert, generating corrupted representations for contrasting with representations from others, forcing each expert to focus on its target crime pattern. (2)\textit{ Hierarchical Adaptive Loss Re-weighting (HALR)}: This algorithm is introduced to dynamically adjust the loss weight of each predictor during the training phase, alleviating issues related to insufficient training in specific regions.

\subsection{Attentive-Gated Mixture-of-Graph-Experts}
Due to the pattern heterogeneity that exists across different crime categories, only modeling a shared spatial-temporal pattern is not able to comprehensively present the complex spatial-temporal dependencies of each crime category.  To address this issue, the MGEs module is proposed. In this module, two types of spatial-temporal graph learning experts are comprised, including a universal expert for shared crime pattern capturing and multiple category-specific experts for distinctive pattern extraction. Spatial attentive gates are applied to selectively route regions to the universal expert or corresponding category-specific expert.

\subsubsection{Spatial-Temporal Graph Learning Expert}

\begin{figure}[h]
    \centering
    \includegraphics[width=1.0\columnwidth]{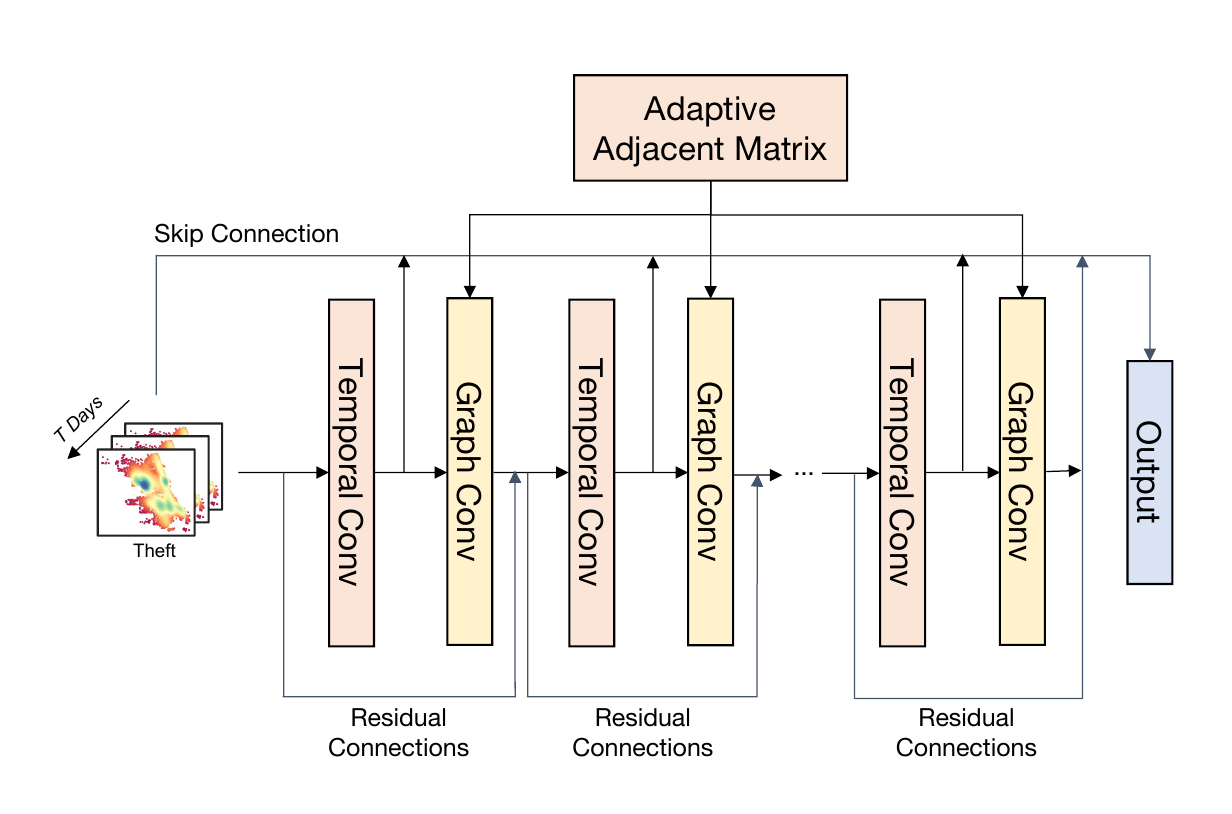}
    \caption{The Structure of Spatial-Temporal Graph Learning Expert}
    \label{fig_expert}
\end{figure} 
The spatial-temporal graph learning expert serves as a key component in MGEs, aiming at modeling both spatial and temporal dependencies simultaneously. The category-specific and universal ST-experts share the same architecture but differ in their data input. The category-specific ST-experts receive crime tensors corresponding to their specific categories, allowing them to model distinctive crime patterns unique to each category. In contrast, the universal ST-expert processes the combined crime tensors from all categories, aiming to capture the shared crime patterns that are common across different types of crimes. Each expert consists of several stacked spatial-temporal blocks (ST-blocks) to simultaneously model the spatial and temporal dependencies across crime occurances. In the ST-blocks, two types of layers are employed: (1) spatial dependency modeling layers, which employ graph convolutional networks (GCN) with self-adaptive adjacency matrices to autonomously capture geographical interrelations, and (2) temporal dependency modeling layers, which utilize dilated temporal convolutional networks (TCN) to capture temporal patterns. The architecture of the ST-expert is illustrated in Figure \ref{fig_expert}.

GCN effectively extends convolutional neural networks to process non-euclidean graph structures. It enhances node representations by aggregating neighbor information with flexible transformation functions, ensuring the retention of structural details. Conventional GCN purely depends on the pre-defined adjacent matrix to provide the structure information, and existing crime prediction approaches construct the adjacent matrix by physical proximity among regions~\cite{xia2021spatial,zhao2023classification}. However, the spatial correlation of crime occurrence not only depends on the physical neighboring but is also influenced by the functions of urban regions. Two regions with similar functions usually present similar crime patterns, such correlations could be ignored in the pre-defined adjacent matrix if they are not physically connected. Consequently, to comprehensively extract spatial dependencies, we adopt a self-adaptive adjacency into the spatial modeling layer, which is entirely learned end-to-end via stochastic gradient descent in the training stage. In particular, for each ST-expert, we define two randomly initialized node embeddings as trainable parameters, $\bm{E}_1, \bm{E}_2 \in \mathbb{R}^{N \times d_n}$, where $d_n$ indicates the hidden size of them. Then the self-adaptive adjacent matrix is constructed as:
\begin{equation}
    \mathcal{A}_{adp} = \text{SoftMax}(\text{ReLU}(\bm{E}_1\bm{E}_2^T)).
\end{equation}
 The SoftMax function is adopted to normalize the self-adaptive adjacency matrix. Consequently, the normalized self-adaptive adjacency matrix can be considered as the transition matrix of a hidden diffusion process. By integrating the pre-defined adjacency matrix $\mathcal{A}$ with self-learned hidden graph dependencies, the adaptive GCN layer can be defined as:

\begin{equation}
\bm{H}^{l+1} = GCN(\bm{H}^l,\mathcal{A}_{adp},\mathcal{A})=\sigma(\mathcal{A}_{adp}\bm{H}^l\bm{W}_1 + \mathcal{A}\bm{H}^l\bm{W}_2 ),
\end{equation}
where $\bm{H}^l$ is the output of the previous layer and $\bm{W}_1$, $\bm{W}_2$ represent the projection matrices with trainable parameters. The function $\sigma(\cdot)$ denotes the activation function, which is set to ReLU in this study. This approach enables the model to uncover implicit dependencies among regions as a complement to the prior spatial information.

TCNs have demonstrated considerable effectiveness in modeling temporal dependencies in spatial-temporal prediction tasks ~\cite{wu2020connecting,yu2018spatio}. Compared to RNNs, TCNs are both time-efficient and parameter-efficient, making them more suitable for urban scenarios where models need to capture features quickly and be sensitive on a limited time scale. We utilize dilated causal convolution to capture the temporal dynamics of crime occurrences. Dilated causal convolution ensures the preservation of temporal causality by introducing zero-padding to the input, thereby ensuring that current predictions are solely based on historical data. The dilated causal convolution operation $F$ on the sequence of one node is defined as:

\begin{equation}
    F(s) = (x*_df)(s) = \sum_{k=0}^{K_t-1}f_k \cdot x_{s-d\cdot k},
\end{equation}

where $d$ is the dilation factor. In the TCN layer, all nodes in the graph can be computed in parallel at the temporal dimension, accelerate. The TCN layer can be expressed as:
\begin{equation}
    \bm{H}^{l+1} = TCN(\bm{H}^l)=\sigma(\text{BN}(F(\bm{H})+\bm{b})),
\end{equation}
where $\bm{H}^l$ is the output of the previous layer and $\bm{b}$ represents the bias. BN indicates the batch normalization layer which normalizes features to accelerate the convergence of the model. $\sigma(\cdot)$ represents the ReLU function.

 Drawing inspiration from ~\cite{wu2020connecting}, in each ST-expert, to mitigate the issue of gradient vanishing, we incorporate residual connections for each spatial-temporal modeling layer, as well as skip connections after each temporal modeling layer.

The representations output by the category-specific ST-expert and the universal ST-expert are denoted as $\bm{H}^{(S)}$ and $\bm{H}^{(U)}$, respectively, encoding the distinctive crime patterns and shared crime patterns accordingly.
 
\begin{figure}[t]
    \centering
    \includegraphics[width=1.0\columnwidth]{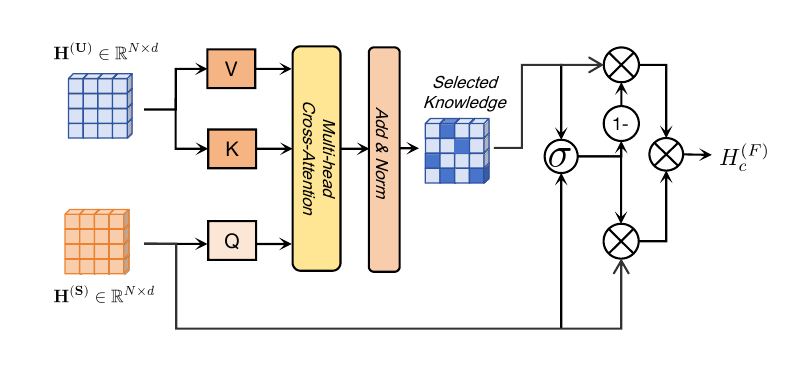}
    \caption{The Structure of the attentive spatial gate}
    \label{fig_gate}
\end{figure} 

\subsubsection{Attentive Spatial Gates}
For different crime categories, there exists a substantial discrepancy in spatial-temporal patterns across regions. This discrepancy indicates that the need for shared knowledge from the universal expert can vary significantly by region for different crime categories. Simply concatenating the knowledge from category-specific experts and the universal expert may introduce noise or even mutually contradictory knowledge in some regions, complicating the accurate modeling of spatial-temporal patterns. To address this issue, we introduce attentive spatial gates for selective knowledge integration from diverse experts.

In particular, we employ a multi-head cross-attention mechanism, which autonomously identifies the relative significance of each region's representations from the universal expert. Specifically, for category $c$ of crime, given the representation $H_{c}^{(S)} \in \mathbb{R}^{N \times d}$ from its category-specific expert, alongside the representation $H^{(U)} \in \mathbb{R}^{N \times d}$ from the universal expert, the multi-head attention mechanism conducts selectively aggregation with attention scores derived from them. The computation for the $m$-th attention head is articulated as:

\begin{equation}
\hat{\bm{H}}_{c}^{(U),m} = \text{softmax}\left(\frac{\bm{W}_Q \cdot \bm{H}_{c}^{(S)} \left( \bm{W}_k \cdot \bm{H}^{(U)} \right)^T}{\sqrt{d_Q}}\right) \bm{W}_V \cdot \bm{H}^{(U)},
\end{equation}

where the matrices $\bm{W}_Q$, $\bm{W}_K$, and $\bm{W}_V$ serve as trainable parameters for queries, keys, and values in the attention model, respectively. The term $d_Q$ denotes the embedding dimension of $\bm{W}_Q$, and the normalization term $\sqrt{d_Q}$ is applied to mitigate overly large inner product values. The representations from each attention head are then merged, described as:

\begin{equation}
\hat{\bm{H}}_{c}^{(U)} = \bm{W^*}\cdot \text{concat}(\hat{\bm{H}}_{c}^{(U),1},...,\hat{\bm{H}}_{c}^{(U),m},...,\hat{\bm{H}}_{c}^{(U),M}),
\end{equation}

where $\bm{W^*}$ is a trainable parameter that combines the outputs of multiple attention heads into a single representation. A gated fusion operation is further applied to adaptively integrate the recalibrated universal representation $\hat{\bm{H}}_{c}^{(U)}$ with the category-specific representation $H_{c}^{(S)}$ in the following manner:

\begin{equation}
\bm{H}^{(F)}_c = \bm{z} \odot \bm{H}_{c}^{(S)} + (1-\bm{z}) \odot \hat{\bm{H}}_{c}^{(U)}
\end{equation}

\begin{equation}
\bm{z} = \sigma(\bm{H}_{c}^{(S)}\bm{W}_{1} + \hat{\bm{H}}_{c}^{(U)}\bm{W}_{2} + \bm{b}_z ),
\end{equation}

where $\bm{W}_{1}$, $\bm{W}_{2}$, and $\bm{b}_z$ denote the trainable parameters of the model. The operation $\odot$ indicates element-wise multiplication, with $\sigma(\cdot)$ being the sigmoid function, acting as a gating mechanism to modulate the integration from each expert. 

\subsubsection{Regional-aware Predictor}
Existing crime prediction approaches usually employ a predictor with shared parameters to predict crime across all regions. However, the skewed crime distribution can introduce substantial bias if a single shared predictor is utilized. Therefore, it is essential to learn a predictor with a separate set of parameters for regions and crime categories with different spatial-temporal characteristics.

To achieve this, we leverage a clustering-based mechanism to encourage regions with similar spatial-temporal characteristics to form clusters. The node embedding $E$ learned in each category-specific ST-expert preserves the unique parameters of each region and reflects their spatial-temporal characteristics. For each crime category, we cluster the regions into $K$ clusters with the K-means algorithm~\cite{krishna1999genetic}. Consequently, the corresponding representations of each cluster are deployed with a Multi-Layer Perceptron (MLP) as the tailored predictor to obtain the prediction results of the involved regions. Formally, the process can be represented as follows:

\begin{equation}
\hat{\bm{X}}_{k,c} = MLP^c_k(\bm{H}^{(F)}_{c,k}),
\end{equation}

where $\bm{H}^{(F)}_{c,k}$ represents the selected representation of region cluster $k$, crime category $c$, and $\hat{\bm{X}}_{k,c}$ denotes the prediction result for this cluster. The integrated prediction result is achieved by aggregating the results from each cluster.

\subsection{Training Strategies}
Although the ST-MoGE network is adept at capturing complex spatial-temporal dependencies inherent in criminal incidents, it is negatively influenced by challenges such as cross-expert information blending and the imbalanced distribution of crime data. In response to these challenges and with the aim of augmenting the model capacity, we propose the implementation of the following training strategies: cross-expert contrastive learning (CECL) and hierarchical adaptive Loss re-weighting (HALR).

\subsubsection{Cross-Expert Contrastive Learning}
Though in the MGEs module, multiple experts are employed to capture different types of crime patterns, without any constraints on the differentiation of target patterns, there may exist blendings across experts in the implementation. Such blending could result in non-eligible redundancy or even contradictory knowledge. To mitigate this issue, we introduce cross-expert contrastive learning, which decomposes the extracted distinctive and mutual patterns in a self-supervised method.

The key insight of CECL is to force the representations encoded with different types of crime patterns to be far away from each other in the latent space. However, since the input for different experts is inherently different, directly leveraging the origin representations to construct negative pairs yields minimal benefit. Therefore, an auxiliary ST-expert is deployed for corrupted representation generation. This expert shares parameters with the universal expert. The corrupted representation of  $c$-th crime category, denoted as $\tilde{\bm{H}}_c$, can be generated by inputting crime data of the corresponding category. Our objective is to maximize the distinction between corrupted representations and ordinary representations. Hence, the negative pairs can be constructed as a corrupted representation with representations from the corresponding category-specific expert or the universal expert. Moreover, inspired by SimCSE ~\cite{gao2021simcse}, the representation for positive pairs construction, denoted as ${\tilde{\bm{H}}_c}'$, are generated in a similar approach, with small disturbance by dropout operations. The InfoNCE~\cite{gutmann2010noise} loss function is adopted, aiming to maximize the lower bound on mutual information for positive pairs while minimizing it for negative pairs~\cite{oord2018representation}. The contrastive loss $\mathcal{L}_{cs}$ for category-specific experts and $\mathcal{L}_{cu}$ for the universal expert can be expressed as:

\begin{equation}
\label{con_loss_s}
\mathcal{L}_{cs} = -\frac{1}{C} \sum_{c=1}^{C} \log \frac{\exp(\text{sim}({\tilde{\bm{H}}_c}', \tilde{\bm{H}}_c) / \tau)}{\sum_{c=1}^{C} {\exp(\text{sim}(\bm{H}^{(S)}_c, \tilde{\bm{H}}_c) / \tau)}},
\end{equation}

\begin{equation}
\label{con_loss_u}
\mathcal{L}_{cu} = -\frac{1}{C} \sum_{c=1}^{C} \log \frac{\exp(\text{sim}({\tilde{\bm{H}}_c}', \tilde{\bm{H}}_c) / \tau)}{\sum_{c=1}^{C} {\exp(\text{sim}(\bm{H}^{(U)}, \tilde{\bm{H}}_c) / \tau)}}.
\end{equation}

Here, $\tau$ represents the temperature parameter, and $\text{sim}(\cdot)$ denotes the cosine similarity function. By applying CECL, the extracted knowledge of different experts can be efficiently separated, consequently decomposing the distinctive and shared crime patterns and relieving the blendings among experts.

\subsubsection{Hierarchical Adaptive Loss Re-weighting}
The real-world crime activities normally follow a long-tailed spatial distribution. 
In the training stage, the model is easily insufficiently trained on regions with relatively low crime frequency due to the skewed crime distribution. To address this challenge, we propose a hierarchical adaptive loss re-weighting algorithm to balance the loss of each cluster of regions, with consideration of imbalanced categorical crime distribution. This method adaptively adjusts weights for each region cluster over time by assessing the change in loss rates. The weighting coefficient for category $c$, $\lambda_c(t)$ is defined as follows:
\begin{equation}
\label{eq13}
    \lambda_c(t) = \frac{C \exp(w_c(t-1)/T)}{\sum_{c=1}^C\exp(w_c(t-1)/T)},
\end{equation}
with
\begin{equation}
\label{sub}
    w_c(t-1) = \frac{\mathcal{L} _c(t-1)}{\mathcal{L} _c(t-2)},
\end{equation}
where $w_c(\cdot)$ measures the rate of loss reduction, $t$ indicates the iteration step, and $T$ denotes a tuning parameter that adjusts the smoothness of weight distribution. A larger $T$ value promotes a more equitable weight distribution. The term $\mathcal{L}_c$ signifies the error in predictions for category $c$.

For clusters belonging to the same crime category, we calculate the coefficient employing a similar method by replacing $\mathcal{L}_c$ in Equation \ref{sub} with $\mathcal{L}_{c,k}$, which indicates the prediction error for cluster $k$ within category $c$.  Consequently, the re-weighted loss is depicted as:
\begin{equation}
\label{eq21}
    \mathcal{L}^{(pred)} = \sum_{c=1}^{C}\lambda_c(t)\sum_{k=1}^{K} \lambda_{c,k}(t)||\bm{X}^{T+1}_{c,k} - \hat{\bm{X}}_{c,k}(t) ||_2^2,
\end{equation}

where $\bm{X}^{T+1}_{c,k}$ and $\hat{\bm{X}}_{c,k}$ respectively denote the ground truth and prediction results for cluster $k$ and category $c$.

\subsection{Joint Optimization}

The composite loss function integrates three key elements. The first element is the re-weighted prediction loss $\mathcal{L}^{(pred)}$, calculated in Equation \ref{eq21}. The second element contains the contrastive losses $\mathcal{L}_{cs}$ and $\mathcal{L}_{cu}$, obtained in equation \ref{con_loss_s}\&\ref{con_loss_u}. 

Overall, the ST-MoGE model is trained by jointly optimizing the following objective function:
\begin{equation}
\label{eq23}
\mathcal{L} = \lambda_1\mathcal{L}^{(pred)} + \lambda_2(\mathcal{L}_{cs}+\mathcal{L}_{cu})
\end{equation}
In this formulation, $\lambda_1$ and $\lambda_2$ act as regulatory coefficients optimizing the balance of losses, collectively summing to 1.

\subsection{Model Complexity Analysis}

In this section, we analyze the time complexity of the ST-MoGE framework. For the MGEs module, it takes $O(C\times L_{(S)} \times N^2 \times T \times d)$ time complexity for all spatial modeling layers and spends $O(C\times L_{(T)} \times N\times T \times d)$ for all temporal modeling layers, where $ L_{(S)}$ and $L_{(T)}$ denotes the number of these layers in each ST-expert. For the attentive spatial gates, it takes $O(C\times M \times N^2 \times d)$ time complexity, where $M$ represents the number of heads in the multi-head attention mechanism. In the CECL module, it takes $O(C\times (L_{(S)} \times N^2 + L_{(T)} \times N) \times T \times d)$ to generate the corrupted representation. The HALR module has a small time complexity that can be neglected. Overall, our ST-MoGE framework can achieve comparable model efficiency compared to GNN-based or attention-based spatial-temporal prediction approaches.

\section{Experiments}
\begin{table}
  \caption{Statistic of Experimented Crime Datasets}
  \centering
  \label{stat}
  \renewcommand{\arraystretch}{1.5} 
  \begin{tabular}{cccc}
    \hline
     \multicolumn{4}{c}{\textbf{New York City Crime Dataset}}\\
    \hline
    \textbf{Time Period} & \multicolumn{3}{c}{July 31, 2020 - November 30, 2022}\\
    \hline
    \textbf{Category} & Larceny & Harassment & Assault\\
    \hline
    \textbf{\#Instances} & 373,702 & 185,628 & 179,231\\
    \hline
    \textbf{Category}& Mischief & Indecency & Robbery\\
    \hline
    \textbf{\#Instances} & 108,093 & 41,700 & 37,314\\
    \hline
     \multicolumn{4}{c}{\textbf{Chicago Crime Dataset}}\\
    \hline
    \textbf{Time Period}  & \multicolumn{3}{c}{July 31, 2020 - July 31, 2023}\\
    \hline
    \textbf{Category} & Larceny & Battery & Damage\\
    \hline
    \textbf{\#Instances}& 191,296 & 121,711 & 78,256\\
    \hline
    \textbf{Category} & Assault & Fraud & Weapons\\
    \hline
    \textbf{\#Instances} & 65,048 & 47,775 & 26,775\\
    \hline
  \end{tabular}
\end{table}
\begin{table*}
  \caption{Overall Performance of Crime Prediction on NYC Dataset in terms OF MAE and MAPE}
  \centering
  \label{table_exp_1}
  \resizebox{\textwidth}{!}{
  \begin{tabular}{ccccccccccccccc}
  
    \toprule
    
    \multirow{2.5}{*}{Model} & \multicolumn{2}{c}{Larceny} &\multicolumn{2}{c}{Harassment} &\multicolumn{2}{c}{Assault} &\multicolumn{2}{c}{Mischief } &\multicolumn{2}{c}{Indecency} &\multicolumn{2}{c}{Robbery} & \multicolumn{2}{c}{Overall} \\
    
    \cmidrule(r){2-3}\cmidrule(r){4-5}\cmidrule(r){6-7}\cmidrule(r){8-9}\cmidrule(r){10-11}\cmidrule(r){12-13}\cmidrule(r){14-15}
    \phantom{a}&MAE{\tiny $\downarrow$}&MAPE{\tiny $\downarrow$} & MAE{\tiny $\downarrow$}&MAPE{\tiny $\downarrow$} & MAE{\tiny $\downarrow$}&MAPE{\tiny $\downarrow$} & MAE{\tiny $\downarrow$}&MAPE{\tiny $\downarrow$} & MAE{\tiny $\downarrow$}&MAPE{\tiny $\downarrow$} & MAE{\tiny $\downarrow$}&MAPE{\tiny $\downarrow$} & MAE{\tiny $\downarrow$}&MAPE{\tiny $\downarrow$}\\
    \midrule
\textbf{STGCN}\cite{yu2018spatio}&1.6206&0.6030&0.9722&0.5649&0.9631&0.5495&0.6654&0.5739&0.3428&0.7353&0.3597&0.7216&0.8206&0.6247\\
\textbf{GWN}\cite{yu2018spatio}&1.4901&0.5627&0.9618&0.5851&0.9490&0.5618&0.6541&0.5814&0.3414&0.7697&0.3594&0.7349&0.7926&0.6326\\
\textbf{GMAN}\cite{zheng2020gman}&1.5748&0.5284&0.9577&0.5779&0.9404&0.5828&0.6463&0.5815&0.3390&0.7193&0.3523&0.6734&0.8018&0.6106\\
\textbf{MTGNN}\cite{wu2020connecting}&1.5807&0.5254&0.9758&0.5900&0.9541&0.5763&0.6570&0.5966&0.3350&0.7562&0.3620&0.7277&0.8108&0.6287\\
\textbf{AGCRN}\cite{bai2020adaptive}&1.5844&0.5328&0.9544&0.5592&0.9409&0.5548&0.6314&0.5662&0.3282&0.7117&0.3593&0.7126&0.7998&0.6062\\
\textbf{MAGNN}\cite{chen2023multi}&1.5189&0.5366&0.9921&0.5699&0.9700&0.5563&0.6608&0.5935&0.3553&0.7039&0.3823&0.6594&0.8132&0.6033\\
\textbf{FCSTGNN}\cite{wang2024data}&1.5679&0.5500&0.9689&0.5989&0.9570&0.6027&0.6514&0.6367&0.3387&0.7525&0.3591&0.7510&0.8072&0.6486\\
\textbf{DeepCrime}\cite{huang2018deepcrime}&1.6917&0.5273&0.9511&0.5604&0.9445&0.5723&0.6518&0.5884&0.3469&0.6625&0.3619&0.6709&0.8247&0.5970\\
\textbf{STHSL}\cite{li2022spatial}&1.8386&0.6186&1.0107&0.6417&0.9927&0.6809&0.6637&0.6712&0.3653&0.6740&0.3656&0.6541&0.8728&0.6567\\
\textbf{STSHN}\cite{xia2021spatial}&1.5457&0.5376&1.0232&0.5572&1.0262&0.5789&0.7066&0.5704&0.3802&0.7479&0.4018&0.7424&0.8473&0.6224\\
\textbf{STGCN-MoE}&1.5282 & 0.6295 & 0.9883 & 0.5790 & 0.9711 & 0.5576 & 0.6698 & 0.5848 & 0.3459 & 0.7504 & 0.3619 & 0.7258 & 0.8109 & 0.6379
\\
\textbf{ST-MOE}\cite{li2023st}&1.5453 & 0.5835 & 0.9833 & 0.5817 & 0.9701 & 0.5649 & 0.6609 & 0.5852 & 0.3417 & 0.7483 & 0.3590 & 0.7149 & 0.8100 & 0.6297
\\
\textbf{ST-MoGE (ours)}&\textbf{1.4516}&\textbf{0.5223}&\textbf{0.9380}&\textbf{0.5529}&\textbf{0.9206}&\textbf{0.5301}&\textbf{0.6196}&\textbf{0.5524}&\textbf{0.3214}&\textbf{0.6399}&\textbf{0.3453}&\textbf{0.6389}&\textbf{0.7661}&\textbf{0.5728}\\

    \midrule
  \end{tabular}
  }
\end{table*}
\begin{table*}
  \caption{Overall Performance of Crime Prediction on CHI Dataset in terms OF MAE and MAPE}
  \centering
  \label{table_exp_2}
  \resizebox{\textwidth}{!}{
  \begin{tabular}{ccccccccccccccc}

    \toprule
    
    \multirow{2.5}{*}{Model} & \multicolumn{2}{c}{Larceny} &\multicolumn{2}{c}{Battery} &\multicolumn{2}{c}{Damage} &\multicolumn{2}{c}{Assault } &\multicolumn{2}{c}{Fraud} &\multicolumn{2}{c}{Weapons} & \multicolumn{2}{c}{Overall} \\
    \cmidrule(r){2-3}\cmidrule(r){4-5}\cmidrule(r){6-7}\cmidrule(r){8-9}\cmidrule(r){10-11}\cmidrule(r){12-13}\cmidrule(r){14-15}
    \phantom{a}&MAE{\tiny $\downarrow$}&MAPE{\tiny $\downarrow$} & MAE{\tiny $\downarrow$}&MAPE{\tiny $\downarrow$} & MAE{\tiny $\downarrow$}&MAPE{\tiny $\downarrow$} & MAE{\tiny $\downarrow$}&MAPE{\tiny $\downarrow$} & MAE{\tiny $\downarrow$}&MAPE{\tiny $\downarrow$} & MAE{\tiny $\downarrow$}&MAPE{\tiny $\downarrow$} & MAE{\tiny $\downarrow$}&MAPE{\tiny $\downarrow$}\\
    \midrule
\textbf{STGCN}\cite{yu2018spatio}&1.2129&0.5634&0.8199&0.5318&0.6698&0.5430&0.5798&0.5106&0.3862&0.7946&0.2935&0.7444&0.6604&0.6146\\
\textbf{GWN}\cite{yu2018spatio}&1.2522&0.5891&0.8239&0.5598&0.6663&0.5336&0.5835&0.5148&0.4036&0.7293&0.2906&0.8083&0.6700&0.6225\\
\textbf{GMAN}\cite{zheng2020gman}&1.2247&0.5516&0.8189&0.5219&0.6645&0.5237&0.5740&0.5120&0.3917&0.8122&0.2771&0.7277&0.6585&0.6082\\
\textbf{MTGNN}\cite{wu2020connecting}&1.2770&0.5715&0.8299&0.5530&0.6693&0.5266&0.5831&0.5402&0.3999&0.7833&0.2913&0.7990&0.6751&0.6289\\
\textbf{AGCRN}\cite{bai2020adaptive}&1.3378&0.5441&0.8284&0.5208&0.6736&0.5140&0.5804&0.5607&0.4091&0.7066&0.2786&0.7310&0.6846&0.5905\\
\textbf{MAGNN}\cite{chen2023multi}&1.2554&0.5624&0.8333&0.5495&0.6731&0.5216&0.5900&0.5102&0.4006&0.7149&0.2904&0.7795&0.6738&0.6064\\
\textbf{FCSTGNN}\cite{wang2024data}&1.2477&0.5661&0.8237&0.5456&0.6755&0.5547&0.5871&0.6131&0.4007&0.8307&0.2826&0.7771&0.6696&0.6479\\
\textbf{DeepCrime}\cite{huang2018deepcrime}&1.2029&0.5470&0.8082&0.5334&0.6670&0.5280&0.5757&0.5403&0.4128&0.7530&0.2846&0.7267&0.6585&0.6047\\
\textbf{STHSL}\cite{li2022spatial}&1.3456&0.5607&0.8412&0.5694&0.6971&0.6030&0.6085&0.5370&0.4074&0.7236&0.3117&0.7490&0.7019&0.6237\\
\textbf{STSHN}\cite{xia2021spatial}&1.3133&0.5983&0.9011&0.5271&0.8157&0.7135&0.6310&0.5493&0.4232&0.8213&0.3092&0.9426&0.7323&0.6920\\
\textbf{STGCN-MoE}&1.2394 & 0.6154 & 0.8217 & 0.5507 & 0.6784 & 0.5380 & 0.5743 & 0.5171 & 0.4045 & 0.7238 & 0.3067 & 0.7287 & 0.6692 & 0.6123
\\
\textbf{ST-MOE}\cite{li2023st}&1.2287 & 0.5770 & 0.8278 & 0.5435 & 0.6730 & 0.5290 & 0.5871 & 0.5420 & 0.4018 & 0.7301 & 0.3041 & 0.8050 & 0.6704 & 0.6211
\\
\textbf{ST-MoGE (ours)}&\textbf{1.1816}&\textbf{0.5357}&\textbf{0.8010}&\textbf{0.5174}&\textbf{0.6571}&\textbf{0.5108}&\textbf{0.5589}&\textbf{0.5076}&\textbf{0.3786}&\textbf{0.7038}&\textbf{0.2731}&\textbf{0.7237}&\textbf{0.6417}&\textbf{0.5812}\\
    \midrule
  \end{tabular}
  }
\end{table*}
\label{sec_exp}

In this section, we conduct comprehensive evaluations of our ST-MoGE framework through extensive experiments on real-world crime datasets, aiming to address the following research questions (RQs):

$\bullet$ \textbf{RQ1}: How does our ST-MoGE framework perform on various crime categories compared to other spatial-temporal prediction models and crime prediction approaches?

$\bullet$ \textbf{RQ2}: What is the impact of our proposed modules, such as MGEs, CECL, and HALR, on the prediction performance?

$\bullet$ \textbf{RQ3}: How do different hyper-parameter settings affect the prediction performance of ST-MoGE?

$\bullet$ \textbf{RQ4}: How does the ST-MoGE framework perform in low-crime-frequency regions?

$\bullet$ \textbf{RQ5}: Is ST-MoGE always effective for different crime categories with heterogeneous spatial-temporal distributions?

\subsection{Experiment Setup}
\subsubsection{Datasets} We evaluate our proposed ST-MoGE framework in two real-world crime datasets, collected from New York City and Chicago. The statistical description of them is presented in Table \ref{stat}. NYC and Chicago are evenly divided into 225 and 140 disjoint geographical regions, respectively. The adjacency
matrix of these regions is constructed based on the geographical proximity relationship between regions. The time slot is set to be 1 day. $T$ is set to 7, which means that data from 7 days before the target day is used as the input to the model. The training, validation, and test data sets are generated by partitioning on both datasets with a ratio of 8:1:1.

\subsubsection{Evaluation Metrics}
To assess the performance of crime prediction, we employ the Mean Absolute Error (MAE) and Mean Absolute Percentage Error (MAPE) as the evaluation metric. MAE measures the average magnitude of errors between actual and predicted values. MAPE calculates the average percentage difference between actual and predicted values, providing a relative measure of accuracy that is easy to interpret. These metrics are defined as follows:
\begin{equation}
    \text{MAE} = \frac{1}{n} \sum_{i=1}^{n} |y_i - \hat{y}_i| ,
\end{equation}
\begin{equation}
\text{MAPE} = \frac{1}{n} \sum_{i=1}^{n} \left| \frac{y_i - \hat{y}_i}{y_i} \right|, 
\end{equation}

where \(y_i\) represents the actual value, \(\hat{y}_i\) represents the predicted value, and \(n\) is the total number of samples.

\subsubsection{Hyper-Parameter Settings}
The hyper-parameters were determined by their performance on the validation set. For each component within the spatial-temporal experts (i.e.: GCN, TCN), the embedding layer, and the attentive spatial gates, the dimensionality of hidden channels is standardized at 32. Within the GCNs, the dimensions of node vectors $E_1$ and $E_2$ were configured to 16, and the kernel size in TCNs is set to 3. In the regional-aware predictor, the number of clusters $K$ is set to be 4. In the context of the hierarchical adaptive loss re-weighting algorithm the temperature parameter $T$ is set to 1, while in the contrastive learning module, the temperature $\tau$ is 0.05. The Adam optimizer was employed for the training process, with an initial learning rate of 0.01. This rate was designed to decrease progressively throughout the training duration. The batch size was established at 64.

\subsubsection{Baselines}
We compare our proposed ST-MoGE framework with 12 baselines for crime prediction. These baseline approaches are categorized into the following three groups:

\textbf{GNN-based Spatial-Temporal Prediction Methods}  
These methods utilize GNNs to capture spatial dependencies. We include STGCN, GWN, GMAN, MTGNN, AGCRN, MAGNN, and FCSTGNN in our comparison.

$\bullet$ \textbf{STGCN}~\cite{yu2018spatio} This method integrates graph convolution for spatial patterns and temporal CNNs for time patterns, enabling comprehensive learning of spatial-temporal relationships in graph-structured data.  

$\bullet$ \textbf{GWN}~\cite{yu2018spatio} This method utilizes an adaptive adjacency matrix in the graph convolutional network to automatically capture the uncertain spatial dependencies, combined with dilated causal convolutions for temporal patterns modeling.  

$\bullet$ \textbf{GMAN}~\cite{zheng2020gman} This method leverages a graph-based attention mechanism to seamlessly integrate spatial and temporal dependencies for comprehensive information aggregation.  

$\bullet$ \textbf{MTGNN}~\cite{wu2020connecting} This method incorporates a graph structure learning module to autonomously capture the latent dependencies while employing graph and temporal convolutional networks to model the spatial and temporal patterns.  

$\bullet$ \textbf{AGCRN}~\cite{bai2020adaptive} This method seamlessly captures node-specific spatial and temporal dependencies through adaptive node embeddings, integrating recurrent neural networks for temporal and adaptive GCNs for spatial dependencies.  

$\bullet$ \textbf{MAGNN}~\cite{chen2023multi} This method exploits a multi-scale network to capture temporal dependencies under different scales of time granularity while using adaptive GCNs to model spatial dependencies at each scale.  

$\bullet$ \textbf{FCSTGNN}~\cite{wang2024data} This method constructs fully connected spatial-temporal graphs to capture dependencies using decay graphs to connect nodes at each time step.

\textbf{Crime Prediction Methods}  
These methods are tailored for crime prediction, trying to tackle specific challenges, such as data sparsity and cross-category dependencies.

$\bullet$ \textbf{DeepCrime}~\cite{huang2018deepcrime} This method leverages multiple auxiliary data sources (e.g., POI, public-service complaints) to enhance prediction accuracy by formulating spatial embeddings and using hierarchical RNNs to capture temporal dependencies.

$\bullet$ \textbf{ST-HSL}~\cite{li2022spatial} This method combines hypergraph-enhanced spatial-temporal convolutional networks for pattern modeling, adopts self-supervised learning for crime data augmentation, addressing data sparsity for robust prediction.  

$\bullet$ \textbf{STSHN}~\cite{xia2021spatial} This method utilizes hypergraph connections for spatial message passing and attention mechanisms to capture evolving temporal relationships, offering an effective framework for spatial-temporal crime data analysis.

\textbf{Mixture-of-Experts Methods}  
To compare MoE architectures, we selected ST-MoE and designed the STGCN-MoE model for this comparison.

$\bullet$ \textbf{STGCN-MoE} This method employs a mixture-of-experts architecture, where multiple spatial-temporal graph convolution networks act as experts, and a fully connected layer serves as the gate for expert selection.

$\bullet$ \textbf{ST-MoE}~\cite{li2023st} This approach stacks convolution-based networks to learn spatio-temporal representations of individual regions and adaptively
assigns appropriate expert layers to different patterns through a spatio-temporal gating network.

\begin{table}
\setlength\tabcolsep{2pt}
  \caption{Module Ablation Study on the Spatial-Temporal Graph Learning Mixture-of-Experts Framework}
  \centering
  \label{ablation}
  \resizebox{\columnwidth}{!}{
  \begin{tabular}{c|cccccc}
    \hline
    \multicolumn{7}{c}{Ablation Study on NYC Dataset}\\
    \hline
     Model & Larceny & Harassment & Assault & Criminal & Indecency & Robbery \\
    \hline
    \textbf{w/o U-Experts} &1.4780&0.9423&0.9256&0.6340&0.3448&0.3522\\
    \textbf{w/o S-Expert} &1.5028&0.9585&0.9410&0.6234&0.3223&0.3472\\
    \textbf{w/o CECL} &1.4648&0.9428&0.9262&0.6346&0.3228&0.3493\\
    \textbf{w/o HALR} &1.4823&0.9411&0.9263&0.6223&0.3256&0.3536\\
    \textbf{ST-MoGE} & \textbf{1.4516}&\textbf{0.9380}&\textbf{0.9206}&\textbf{0.6196}& \textbf{0.3214} &\textbf{0.3453}\\
    \hline
    \hline
     \multicolumn{7}{c}{Ablation Study on CHI Dataset}\\
    \hline
    Model & Larceny & Battery & Damage & Assault & Fraud & Weapons\\
    \hline
    \textbf{w/o U-Experts} &1.2046&0.8168&0.6623&0.5689&0.3896&0.2820\\
    \textbf{w/o S-Expert} &1.2283&0.8265&0.6662&0.5749&0.3795&0.2752\\
    \textbf{w/o CECL} &1.2018&0.8089&0.6592&0.5599&0.3837&0.2755\\
    \textbf{w/o HALR} &1.1979&0.8172&0.6674&0.5742&0.3807&0.2850\\
    \textbf{ST-MoGE}&\textbf{1.1816}&\textbf{0.8010}&\textbf{0.6571}&\textbf{0.5589}& \textbf{0.3786} &\textbf{0.2731}\\

    \hline
    
  \end{tabular}
  }
\end{table}

\begin{figure*}[t]
    \centering
    \includegraphics[width=\textwidth]{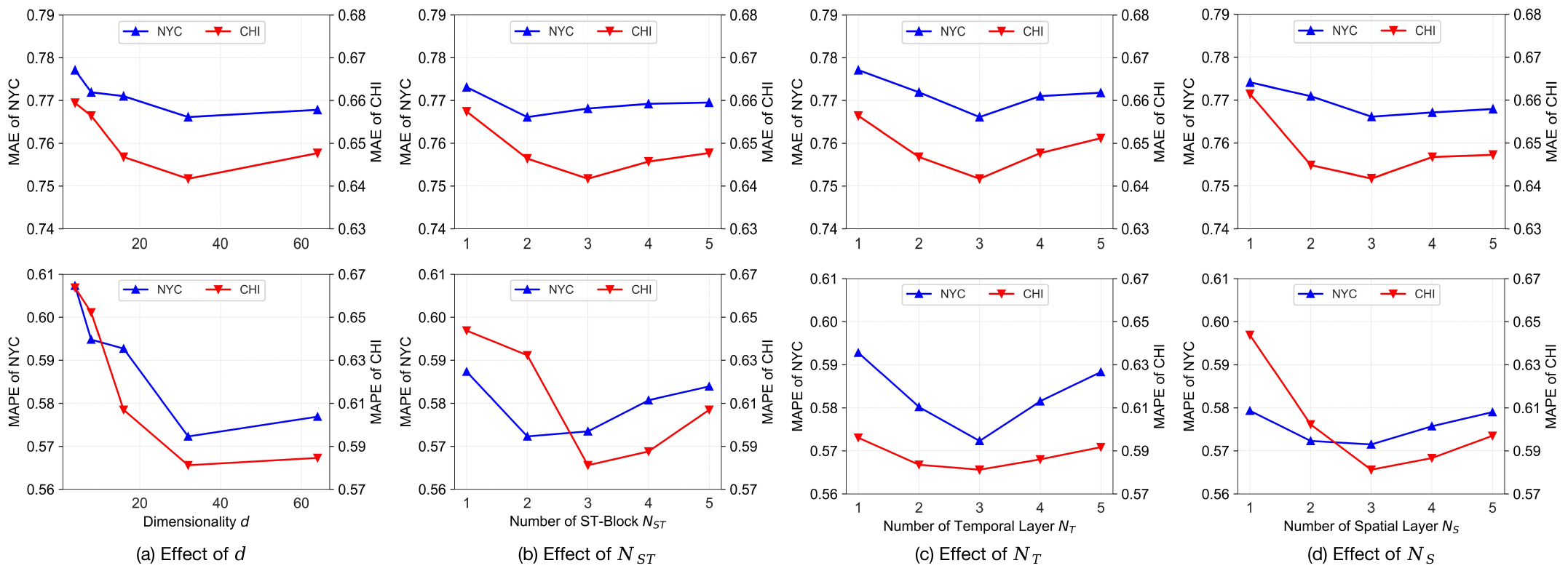}
    \caption{Impact Study for Hyperparameters on Chicago and New York Crime Data}
    \label{rob}
\end{figure*}

\subsection{Performance Comparison (RQ1)}
Table \ref{table_exp_1} and \ref{table_exp_2} report the performance comparison results of between our ST-MoGE framework and various baseline models for crime prediction. We summarize our findings as follows:

$\bullet$ ST-MoGE outperforms all the compared baseline models on all crime categories within both datasets. We attribute such improvements to: {\romannumeral1}) By leveraging MoE architecture network, ST-MoGE could comprehensively preserve heterogeneous crime patterns. {\romannumeral2}) Benefiting from contrastive learning, ST-MoGE is able to decompose different types of crime patterns, reducing the modeling pattern information loss. {\romannumeral3}) With the design of clustered predictors and reweighting strategy, ST-MoGE could better capture the crime patterns on regions with low crime rates under the imbalanced crime distribution. 

$\bullet$ Though all the baseline models can effectively capture spatial-temporal dependencies, their performances are constrained because their architectures are limited to capturing only shared crime patterns. Such limitations make their capacities insufficient to comprehensively preserve the complex and heterogeneous crime patterns of different types. Moreover, the captured crime pattern of these models is usually biased to specific categories while performing relatively worse on other categories. For example, in the CHI crime dataset, STGCN and DeepCrime demonstrate remarkable accuracy on high-frequency categories, such as larceny, but perform poorly on low-frequency categories like robbery. Conversely, AGCRN exhibits better performance on low-frequency categories but struggles with high-frequency ones. This is because mutual contradictions potentially exist between crime patterns of different categories, leading to biases in some categories. Benefiting from the MoE architecture, our ST-MoGE well handled the spatial-temporal heterogeneity, performing remarkable prediction ability on all the crime categories. 

\begin{figure*}[t]
    \centering
    \includegraphics[width=1.0\linewidth]{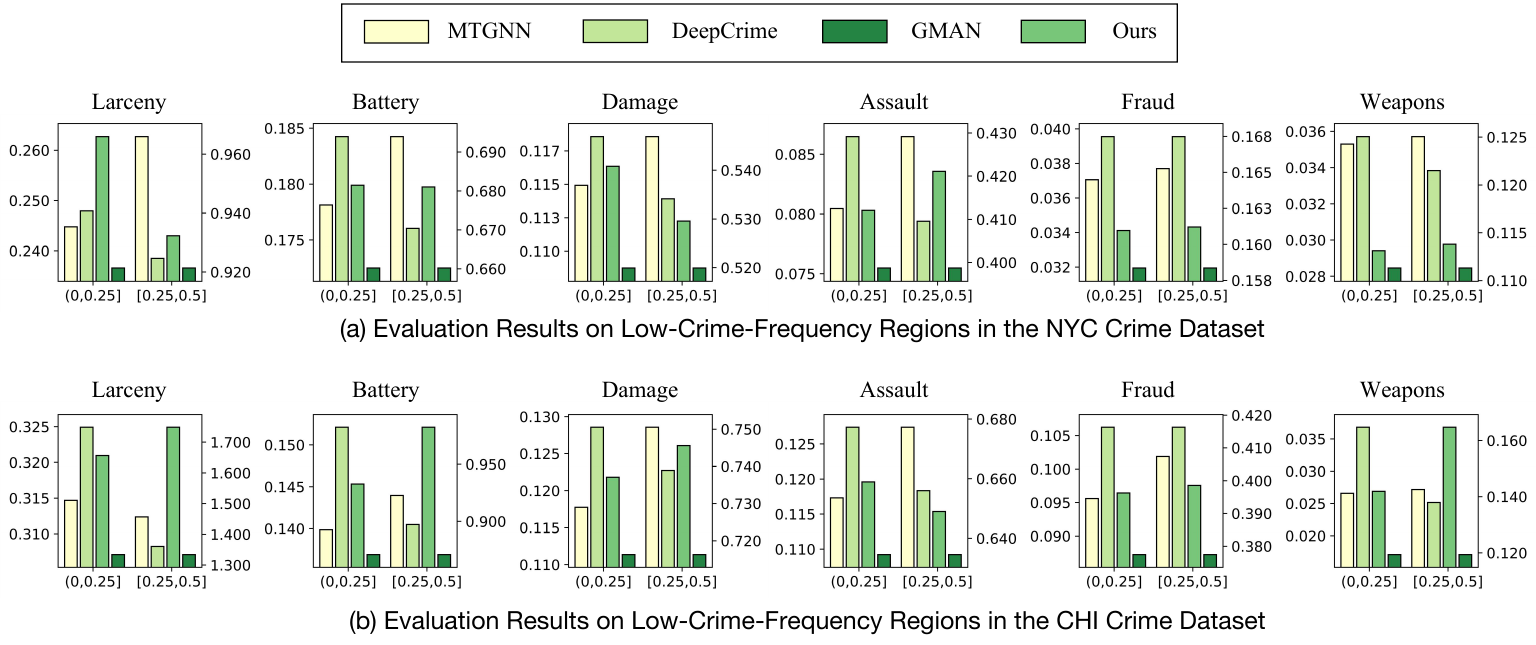}
    \caption{Investigation on the effectiveness of ST-MoGE on low-crime-frequency regions in Chicago and New York}
    \label{imbala}
\end{figure*}

\subsection{Ablation Study (RQ2)}

To investigate the effectiveness of different components, we conduct ablation studies on both crime datasets.

\subsubsection{Effect of MGEs}
Compared with conventional spatial-temporal prediction approaches, a significant difference in our method is the adoption of a MoE architecture, which deploys multiple ST-experts to model spatial-temporal dependencies from different aspects. To evaluate the effectiveness of the MoE architecture, we construct two variants, named "w/o S-Experts" and "w/o U-Expert," by disabling the category-specific ST-experts and the universal expert, respectively. The results in Table \ref{ablation} demonstrate that removing each type of expert leads to a reduction in prediction performance. In particular, disabling the category-specific experts significantly impacts the prediction capability for the majority category (e.g., Larceny). Without the category-specific experts, the MoE network degenerates into a conventional spatial-temporal graph neural network where all categories share one spatial-temporal pattern, adversely affected by spatial-temporal heterogeneity. Conversely, the removal of the universal expert reduces prediction accuracy for minority categories (e.g., Indecency, Robbery in the NYC dataset; Fraud, Weapons in the CHI dataset) due to severe sparsity in crime data for these categories, which challenges effective pattern extraction. These experiments underscore the benefits of the MoE architecture, where each type of expert complements the other, positively impacting overall crime prediction performance.

\subsubsection{Effect of CECL}
We further conduct experiments by removing the CECL module to evaluate its effectiveness, naming this variant "w/o CECL." The experiment results are presented in Table \ref{ablation}. We observe that the removal of CECL leads to performance drops across all categories. This demonstrates the effectiveness of the CECL module and underscores the necessity of decomposing different types of crime patterns.

\subsubsection{Effect of HALR}
To verify the effectiveness of our proposed hierarchical adaptive loss re-weighting algorithm (HALR), we conduct experiments on the variant "w/o Reweight," which does not employ the hierarchical adaptive loss re-weighting algorithm during the training phase, setting the loss weights of each region cluster equally. The results presented in Table \ref{ablation} show that removing HALR leads to performance degradation, confirming the effectiveness of HALR.

\begin{figure*}[t]
    \centering
    \includegraphics[width=1.0\linewidth]{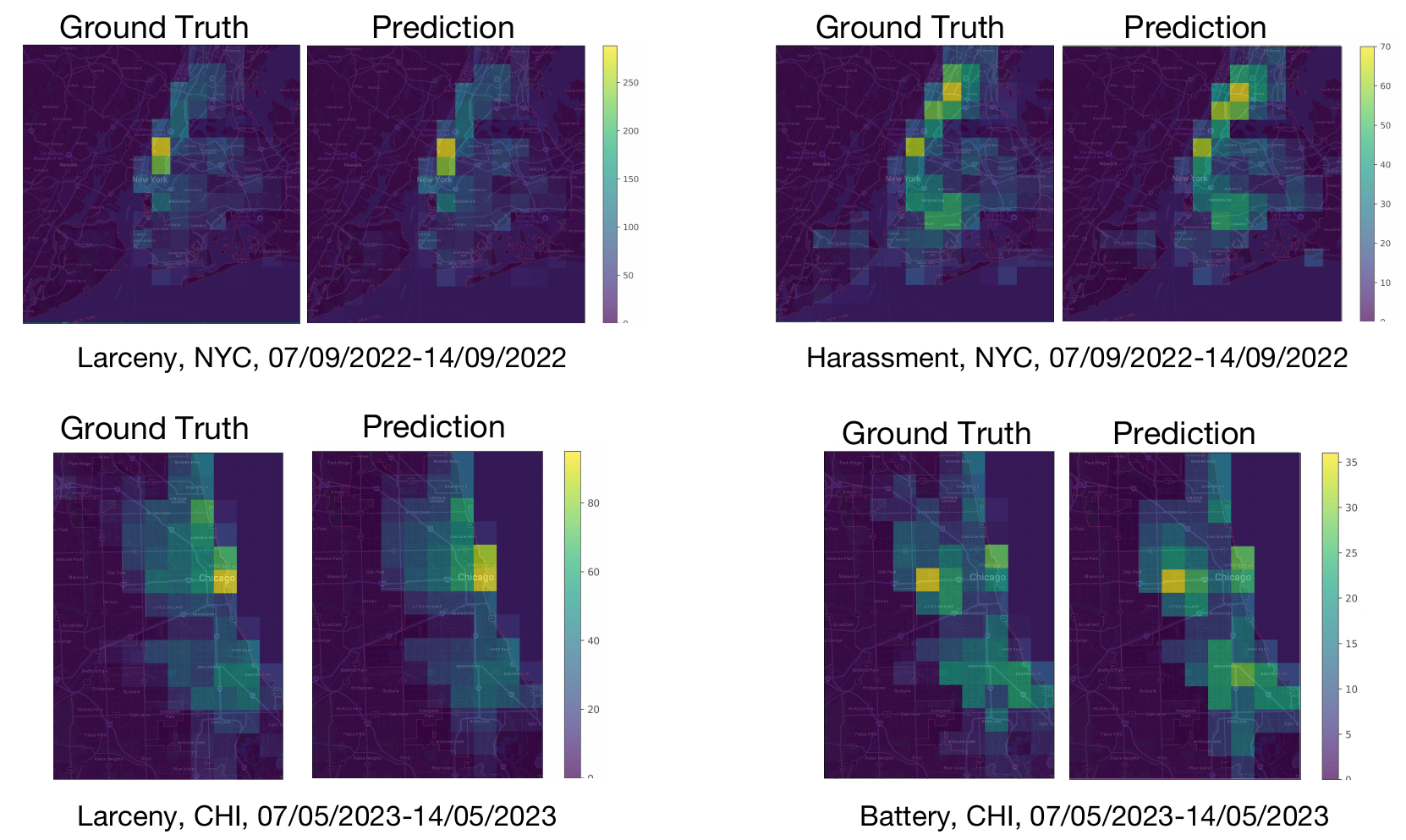}
    \caption{Visualization for Prediction Results on Diverse Crime Categories on Chicago and New York Crime Data}
    \label{vis}
\end{figure*}
\subsection{Hyperparameter Study (RQ3)}

In this section, we conduct a series of experiments to explore the influence of different hyperparameter settings on our framework's performance. We summarize the observations below to analyze the influence of different settings:

\subsubsection{Dimensionality $d$}
We explore the representation dimensionality in the range of $\{2^2, 2^3, 2^4, 2^5, 2^6\}$. The experiment results are shown in Figure \ref{rob} (a), demonstrating that when the dimensionality $d=32$, the overall prediction accuracy approaches the best. If the dimensionality is too small, it could limit the network's representation ability and lead to underfitting. Conversely, increasing the dimensionality too much leads to a slight reduction in prediction performance, likely due to the issue of overfitting.

\subsubsection{Depth of Experts $N_{ST}$}
We examine the number of ST-blocks in each ST-expert within the range $\{1, 2, 3, 4, 5\}$, and the results are shown in Figure \ref{rob} (b). The results show that experts with 3 ST-blocks outperform others in both datasets. While increasing model depth can enhance representation ability, stacking too many layers may lead to overfitting.

\subsubsection{Number of Spatial Modeling Layers $N_{S}$}
We test the influence of varying the number of spatial modeling layers in each ST-block in the range of $\{1, 2, 3, 4, 5\}$. The results, presented in Figure \ref{rob} (c), show that 2 layers perform the best in the NYC dataset, while 3 layers perform the best in the CHI dataset. As the number of GNN layers increases, the spatial receptive fields of the nodes expand, enhancing the ST-block's ability to process and extract high-dimensional features more effectively. However, increasing the depth too much may lead to over-smoothing.

\subsubsection{Number of Temporal Modeling Layers $N_{T}$}
We test the influence of different numbers of temporal modeling layers in each ST-block within the range of $\{1, 2, 3, 4, 5\}$. The results, presented in Figure \ref{rob} (d), demonstrate that 3 layers perform the best in both datasets. As the number of TCN layers increases, the temporal receptive fields expand, facilitating complex feature extraction. However, excessive depth may involve unexpected noise in representation learning.

\subsection{Effectivenss on Low-frequency Regions (RQ4)}
In this section, we perform experiments to validate the effectiveness of our ST-MoGE in regions with low crime frequency. We separately evaluate the prediction performance of all crime categories in regions with relatively low crime frequency. For each category, we split regions with less frequent crimes into two groups, at the quantile intervals $(0.0,0.25]$ and $(0.25,0.5]$. The evaluation results are presented in Figure \ref{imbala}.

We observe that our ST-MoGE model outperforms almost all other methods in all cases, demonstrating the effectiveness of ST-MoGE in regions with extreme situations. The imbalanced spatial distribution of crime occurrences negatively affects spatial-temporal dependencies modeling for neural networks, often leading to biases and insufficient training on low-frequency regions, making it difficult to effectively learn crime patterns in these areas. With the incorporation of our HALR module, the ST-MoGE can automatically adjust the importance of regions during the training stage, alleviating the impacts of imbalanced distribution.
\begin{figure}[h]
    \centering
    \includegraphics[width=1.0\columnwidth]{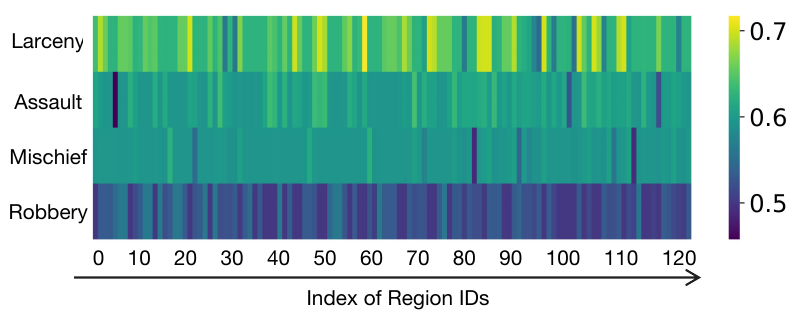}
    \caption{Gate Weights of Each Region and Different Crime Categories. }
    \label{gate}
\end{figure}

\subsection{Case Study}
To intuitively present the ability of our ST-MoGE in crime prediction, we visualize the prediction results alongside the ground truth for the two datasets. Figure \ref{vis} shows the visualization results. For each dataset, we select two crime categories with notable distribution heterogeneity.

Overall, the predicted heat maps of crimes on both datasets closely resemble the ground truth, indicating that our model provides highly accurate predictions. Notably, the visualizations reveal that crime occurrences are significantly imbalanced; most regions exhibit a very small number of crimes, while only a small proportion of regions (primarily central urban areas) experience high crime rates. Despite this imbalance, the prediction results for regions with low crime rates are still accurate, demonstrating the robustness of the ST-MoGE's performance on imbalanced data distributions.

To enrich the comprehension of our knowledge selection process, we further present a qualitative example that illustrates its details. Figure \ref{gate} illustrates the weights of the attentive spatial gates for each region and different crime categories on May 28, 2023, in New York City. From this figure, it is evident that the weights vary significantly across different categories within each region, indicating that the attentive spatial gates are effectively selecting the necessary knowledge from the shared ST-expert tailored to each crime category. This differentiation in weights highlights the attentive spatial gates' ability to adaptively integrate relevant spatial-temporal knowledge, thereby optimizing the prediction capacity for diverse crime types.

\section{Conclusion}
\label{sec_con}

In this paper, we present an effective Spatial-Temporal Mixture-of-Graph-Experts (ST-MoGE) framework to address the crime prediction problem. Our approach introduces an Attentive-Gated Mixture-of-Graph-Experts (MGEs) module, which comprehensively captures the heterogeneous spatial-temporal dependencies of crime occurrences. Additionally, we incorporate a Cross-Expert Contrastive Learning (CECL) paradigm to decompose different types of crime patterns, enhancing expert focus and alleviating mutual contradictions among experts. To further improve performance, we employ clustered predictors and a Hierarchical Adaptive Loss Re-weighting (HALR) scheme, which mitigates modeling biases and ensures sufficient training for regions with extreme crime frequencies. When evaluated on real-world datasets, ST-MoGE outperformed existing state-of-the-art methods and demonstrated the efficacy of each key component. Although specifically designed for crime prediction, the foundational principles of our model have broader applications and could be beneficial in other areas of multi-objective spatial-temporal prediction, such as urban anomalies prediction.


\addcontentsline{toc}{chapter}{Bibliography}
\bibliographystyle{IEEEtran}
\bibliography{references}

\vfill

\end{document}